\documentclass{article}
\usepackage{spconf,amsmath,graphicx}
\usepackage[font=small,skip=1pt]{caption}
\usepackage{comment}
\usepackage{amsmath}
\usepackage{amssymb}
\usepackage{caption}
\usepackage{subcaption}
\usepackage[linesnumbered,ruled,vlined]{algorithm2e}
\usepackage{comment,color,soul}
\usepackage{color, colortbl}
\usepackage{subcaption}
\usepackage{mathtools}
\usepackage{amsfonts}       %
\usepackage{amsmath}        %
\usepackage{mathtools}      %
\usepackage{multirow}
\usepackage{nicefrac}       %
\usepackage{microtype} 
\usepackage{hyperref}

\title{A Tunable Robust Pruning Framework\\ Through \underline{D}ynamic \underline{N}etwork \underline{R}ewiring of DNNs}
\name{Souvik Kundu , Mahdi Nazemi, Peter A. Beerel, Massoud Pedram
\thanks{This work was partly supported by NSF including grant \#1763747.}
\thanks{Accepted as a conference paper at ASP-DAC, 2021.}}
\address{University of Southern California, Los Angeles, CA 90089}
\begin{document}
%
\maketitle
\begin{abstract}
This paper presents a dynamic network rewiring (DNR) method to generate pruned deep neural network (DNN) models that are robust against adversarial attacks yet maintain high accuracy on clean images. In particular, the disclosed DNR method is based on a unified constrained optimization formulation using a hybrid loss function that merges ultra-high model compression with robust adversarial training. This  training strategy dynamically adjusts inter-layer connectivity based on per-layer normalized momentum computed from the hybrid loss function. In contrast to existing robust pruning frameworks that require multiple training iterations, the proposed learning strategy achieves an overall target pruning ratio with only a single training iteration and can be tuned to support both irregular and structured channel pruning. To evaluate the merits of DNR, experiments were performed with two widely accepted models, namely VGG16 and ResNet-18, on CIFAR-10, CIFAR-100 as well as with VGG16 on Tiny-ImageNet. Compared to the baseline uncompressed models, DNR provides over $20\times$
compression on all the datasets with no significant drop in either clean or adversarial classification accuracy. Moreover, our experiments show that DNR consistently finds compressed models with better clean and adversarial image classification performance than what is achievable through state-of-the-art alternatives. 
\end{abstract}
\section{Introduction}
\label{sec:intro}
In recent years, deep neural networks (DNNs) have emerged as critical components in various applications, including image classification \cite{krizhevsky2012imagenet}, speech recognition \cite{hinton2012deep}, medical image analysis \cite{litjens2017survey} and autonomous driving \cite{chen2015deepdriving}.
However, despite the proliferation of deep learning-powered applications, machine learning models have raised significant security concerns due to their vulnerability to  {\em adversarial examples}, i.e., maliciously generated images which are perceptually similar to clean ones with the ability to fool classifier models into making wrong predictions \cite{carlini2017towards, goodfellow2014explaining}. Various recent work have proposed associated defense mechanisms including adversarial training \cite{goodfellow2014explaining}, hiding gradients \cite{tramer2017ensemble}, adding noise to the weights \cite{he2019parametric}, and several others \cite{ meng2017magnet}. 
\begin{figure}[!t]
\includegraphics[width=0.98\linewidth]{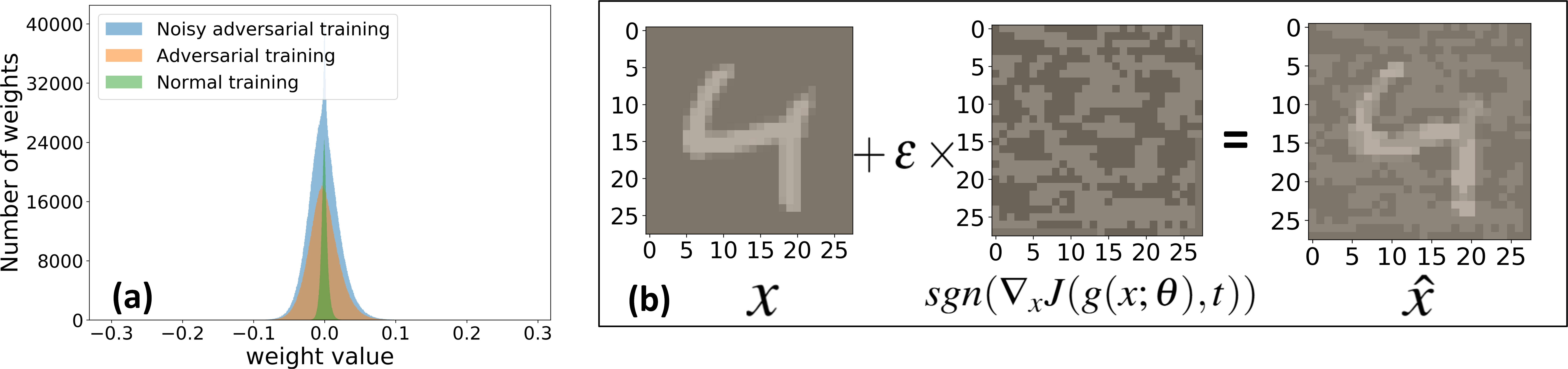}
\centering
\caption{(a) Weight distribution of the $14^{th}$ convolution layer of ResNet18 model for different training schemes: normal, adversarial \cite{madry2017towards}, and noisy adversarial \cite{he2019parametric}. (b) An adversarially generated image ($\hat{\textbf{\em x}}$) obtained through FGSM attack, which is predicted to be the number 5 instead of 4 ($\textbf{\em x}$).}
\label{fig:res18_wt_hist_and_fgsm}
\vspace{-3mm}
\end{figure}

Meanwhile, large model sizes have high inference latency, computation, and storage costs that represent significant challenges in deployment on IoT devices.
Thus reduced-size models \cite{kundu2019psconv, fayyazi2019csrram} and model compression techniques e.g., {\em pruning} \cite{dettmers2019sparse, ding2019global, He_2018_ECCV}, 
have gained significant traction. In particular, earlier work showed that without a significant accuracy drop, pruning can remove more than $90\%$ of the model parameters \cite{dettmers2019sparse, ding2019global} and that ensuring the pruned models have structure can yield observed performance improvements on a broad range of compute platforms \cite{he2017channel}.
However, 
adversarial training that increases network robustness generally demands more non-zero parameters than needed for only clean data \cite{madry2017towards} as illustrated in Fig. \ref{fig:res18_wt_hist_and_fgsm}(a). Thus a naively compressed model performing well on clean images, can become vulnerable to adversarial images.      
Unfortunately, despite a plethora of work on compressed model performance on clean data, there have been only a few studies on the robustness of compressed models under adversarial attacks. 


In particular, some prior works \cite{ye2019adversarial, gui2019model} have tried to 
design a compressed yet robust model through a unified constrained optimization formulation by using the alternating direction method of multipliers (ADMM) in which dynamic $L_2$ regularization is the key to outperforming state of the art pruning techniques \cite{ren2019admm}. However, in ADMM the network designer needs to specify layer-wise sparsity ratios, which requires prior knowledge of an effective compressed model. This knowledge may not be available and thus training may require multiple iterations to determine good layer-sparsity ratios. Another related research \cite{sehwag2020hydra} has aimed to use pre-trained weights to perform robust pruning and has demonstrated the benefits of fine-tuning after training in terms of increased performances.
In other schemes like Lasso \cite{rakin2019robust}, a target compression ratio cannot be set because the final compression ratio is not determined until training is completed. Moreover, Lasso requires separate re-training to increase the accuracy after the  assignment of non-significant weights to zero, resulting in costly training. 

In contrast, this paper presents  dynamic network rewiring (DNR), a unified training framework to find a compressed model with increased robustness that does not require individual per-layer target sparsity ratios. In particular, we introduce a hybrid loss function for robust compression which has three major components: a clean image classification loss, a dynamic $L_2$-regularizer term inspired by a relaxed version of ADMM \cite{dinh2018convergence}, and an adversarial training loss. Inspired by {\em sparse-learning}-based training scheme of \cite{dettmers2019sparse}, we then propose a single-shot training framework to achieve a robust pruned DNN using the proposed loss. In particular, DNR dynamically arranges per layer pruning ratios using normalized momentum, 
maintaining the target pruning every epoch, without requiring any fine tuning.
 In summery, our key contributions are:
\vspace{-0.1cm}
\begin{itemize}
    \item Given only a global pruning ratio, we propose a single-shot (non-iterative) training framework that simultaneously achieves ultra-high compression ratio, state-of-the-art accuracy on clean data, and robustness to perturbed images.
 
    \item We extend the approach to support structured pruning technique, namely {\em channel pruning}, enabling benefits on a broader class of compute platforms. As opposed to conventional sparse-learning \cite{dettmers2019sparse} that can perform only irregular pruning, models generated through structured DNR can significantly speed up inference. To the best of our knowledge, we are the first to propose a non-iterative robust training framework that supports both irregular and channel pruning. 
    \item We provide a comprehensive investigation of adversarial robustness for both channel and irregular pruning, and obtain insightful observations through evaluation on an extensive set of experiments on CIFAR-10 \cite{krizhevsky2009learning}, CIFAR-100  \cite{krizhevsky2009learning}, and Tiny-ImageNet \cite{hansen2015tiny} using variants of ResNet18 \cite{he2016deep} and VGG16 \cite{simonyan2014very}. 
Our proposed method consistently outperforms state-of-the-art (SOTA) \cite{rakin2019robust, ye2019adversarial} approaches with negligible accuracy drop compared to the unpruned baselines.
\end{itemize}

We further empirically demonstrate the superiority of our scheme when used to target model compression on clean-only image classification task compared to SOTA non-iterative pruning mechanisms \cite{dettmers2019sparse, ding2019global, He_2018_ECCV, lee2018snip}.\footnote{This paper targets low-cost training, thus comparisons to iterative pruning methods (e.g., \cite{ren2019admm}) are out of scope.}


The remainder of this paper is structured as follows. In section \ref{sec:back} we present necessary background work. Section \ref{sec:proposed_method} describes proposed DNR based training method. We present our experimental results in Section \ref{sec:res} and conclude in Section \ref{sec:con}.

\section{Background Work} 
\label{sec:back}
\subsection{Adversarial Attacks}

Recently, various adversarial attacks have 
been proposed to find fake images, i,e.,  adversarial examples, which have barely-visible perturbations from real images but still manage to fool a trained DNN. 
One of the most common attacks is the fast gradient sign method (FGSM) \cite{goodfellow2014explaining}. Given a vectorized input \textit{\textbf{x}} of the real image and corresponding label \textit{\textbf{t}}, FGSM perturbs each element \textit{x} in \textit{\textbf{x}} along the sign of 
the associated element of the gradient of the inference loss w.r.t. \textit{\textbf{x}} as shown in Eq. \ref{eq:fgsm} and illustrated in  Fig. \ref{fig:res18_wt_hist_and_fgsm}(b). 
Another common attack is the projected gradient descent (PGD) \cite{madry2017towards}.
The PGD attack is a multi-step variant of FGSM where $\hat{\textbf{\em x}}^{k=1} = {\textbf{\em x}}$ and the iterative update of the perturbed data $\hat{\textbf{\em x}}$ in $k^{th}$ step is given in Eq. \ref{eq:pgd}.   
\begin{align}
\vspace{-0.3cm}
\hat{\textbf{\em x}}&={\textbf{\em x}} + \epsilon \times sgn(\nabla_{x}J(g(\textbf{\em  x}; \pmb{\theta}), \textbf{\em t})) \label{eq:fgsm} \\
\hat{\textbf{\em x}}^{k}&=\mbox{Proj}_{P_{\epsilon}(\textit{\textbf{x}})} (\hat{\textbf{\em x}}^{k-1} + \alpha \times sgn(\nabla_{x}J(g(\hat{\textbf{\em x}}^{k-1}; \pmb{\theta}), \textbf{ \em t}))) \label{eq:pgd}
\end{align}
Here, the scalar $\epsilon$ corresponds to the perturbation constraint that determines the severity of the perturbation. $g(\textbf{\em x}, \pmb{\theta})$ generates the output of the DNN, parameterized by $\pmb{\theta}$. Here, $\mbox{Proj}$ projects the updated adversarial sample onto the projection space $P_{\epsilon}(\textit{\textbf{x}})$  which is the   $\epsilon$-$L_{\infty}$ neighbourhood of the benign sample \footnote{It is noteworthy that the generated $\hat{\textbf{\em x}}$ are clipped to a valid range which for our experiments is $[0,1]$.} $\textit{\textbf{x}}$. $\alpha$ is the attack step size. 

Note that these two strategies assume the attacker knows the details of the DNN and are thus termed as {\em white-box attacks}. We will evaluate the merit of our training scheme by measuring the robustness of our trained models to the fake images generated by these attacks. We argue that this evaluation is more comprehensive than using images generated by attacks that assume limited knowledge of the DNN \cite{ren2020adversarial}. 
Moreover, we note that PGD is one of the strongest $L_{\infty}$ adversarial example generation algorithms \cite{madry2017towards} and use it as part of our proposed framework.
\subsection{Model Compression}
ADMM is a powerful optimization method used to solve problems with non-convex, combinatorial constraints \cite{boyd2011distributed}. 
It decomposes the original optimization problem into two sub-problems and solves the sub-problems iteratively until convergence. 
Pruning convolutional neural networks (CNNs) can be modeled as an optimization problem where the cardinality of each layer's weight tensor is bounded by its pre-specified pruning ratio.
In the ADMM framework, such constraints are transformed to ones represented with indicator functions, such as $\mathbb{I}_{\theta}(\textbf{z})= 0$ for $|\textbf{z}|  \leq n$ and $+\infty $ otherwise.
%
%
%
Here, $\mathbf{z}$ denotes the duplicate variable \cite{boyd2011distributed} and $n$ represents the target number of non-zero weights determined by pre-specified pruning ratios. 
Next, the original optimization problem is reformulated as: 
\begin{equation}
    \mathcal{L}_{\rho}(\pmb{\theta},\textbf{z},\lambda) = J(g(\textbf{\em x};\pmb{\theta}), \textbf{\em t}) + \mathbb{I}_{\theta}(\textbf{z}) + \left\langle  \lambda, \pmb{\theta} - \textbf{z} \right\rangle  + \frac{\rho}{2} ||\pmb{\theta} - \textbf{z}||^2_2
    \label{eq:admm}
\end{equation}
where $\lambda$ is the Lagrangian multiplier and $\rho$ is the penalization factor when parameters $\pmb{\theta}$ and $\mathbf{z}$ differ. 
Eq. \eqref{eq:admm} is broken into two sub-problems which solve $\pmb{\theta}$ and $\mathbf{z}$ iteratively until convergence \cite{ren2019admm}. 
The first sub-problem uses stochastic gradient descent (SGD) to update  $\pmb{\theta}$ while the second sub-problem applies projection to find the assignment of $\mathbf{z}$ that is closest to $\pmb{\theta}$ yet satisfies the cardinality constraint, effectively pruning weights with small magnitudes. 

Not only can ADMM prune a model's weight tensors but it also has as a dynamic regularizer. 
Such adaptive regularization is one of the main reasons behind the success of its use in pruning. 
However, ADMM-based pruning has several drawbacks. 
First, ADMM requires prior knowledge of the per-layer pruning ratios. 
Second, ADMM does not guarantee the pruning ratio will be met, and therefore, an additional round of hard pruning is required after ADMM completes. 
Third, not all problems solved with ADMM are guaranteed to converge. 
Fourth, to improve the convergence, $\rho$ needs to be progressively increased across several rounds of training, which increases training time \cite{boyd2011distributed}. 

Sparse learning \cite{dettmers2019sparse} addresses the shortcomings of ADMM by leveraging exponentially smoothed gradients (momentum) to prune weights. 
It redistributes pruned weights across layers according to their mean momentum contribution. 
The weights that will be removed and transferred to other layers are chosen according to their magnitudes while the weights that are brought back (reactivated) are selected based on their momentum values. 
%
On the other hand, a major shortcoming of sparse learning compared to ADMM is that it does not benefit from a dynamic regularizer and thus often yields lower levels of accuracy. 
Furthermore, existing sparse-learning schemes only support irregular forms of pruning, limiting speed-up on many compute platforms. Finally, sparse-learning, to the best of our knowledge, has not previously been extended to robust model compression.
\section{Dynamic Network Rewiring}
\label{sec:proposed_method}
To tackle the shortcomings of ADMM and sparse-learning this section introduces a dynamic $L_2$ regularizer that enables non-iterative training to achieve high accuracy with compressed models. 
We then describe a hybrid loss function to provide robustness to the compressed models and an extension to support structured pruning. 

\subsection{Dynamic $L_2$ Regularizer}
\label{dyn_l2}

%
%
For a DNN parameterized by $\pmb{\theta}$ with $L$ layers, we let $\pmb{\theta}_l$ represent the weight tensor of layer $l$. In our sparse-learning approach, these weight tensors are element-wise multiplied ($\odot$) by corresponding binary mask tensors ($\mathbf{m}_l$)
to retain only a fraction of non-zero weights, thereby meeting a target pruning ratio. We update each layer mask in every epoch similar to \cite{dettmers2019sparse}. The number of non-zeros is updated based on the layer's normalized momentum and the specific non-zero entries are set to favor large magnitude weights.
We incorporate an ADMM dynamic $L_2$ regularizer \cite{ren2019admm} into this framework by introducing duplicate variable $\mathbf{z}$ for the non-zero weights, which is in turn updated at the start of every epoch.  Unlike \cite{ren2019admm}, we only penalize differences between the masked weights ($\pmb{\theta}_l \odot \textbf{m}_l$) of a layer $l$ and their corresponding duplicate variable $\mathbf{z}_l$.  
Because the total cardinality constraint of the masked parameters is satisfied, i.e. $\sum_{l=1}^L \mbox{card}(\pmb{\theta}_l \odot \mathbf{m}_l) \le n$, the indicator penalty factor is redundant and the loss function may be simplified as
%
%
%
%
%
\vspace*{-0.2cm}
\begin{align}
    {\mathcal{L}}_{\rho}(\pmb{\theta},\textbf{z}, \mathbf{m}) = J(g(\textbf{\em x};\pmb{\theta}, \mathbf{m}), \textbf{\em t}) + 
    \frac{\rho}{2}\sum_{l=1}^{L} ||\pmb{\theta}_l \odot \mathbf{m}_l - \textbf{z}_l||^2_2 
    \label{eq:admm-simplified}
\end{align}
%
where, $\rho$ is the dynamic $L_2$ penalizing factor. This simplification is particularly important because the indicator function used in Eq. \ref{eq:admm} is non-differentiable and its removal in Eq. \ref{eq:admm-simplified} enables the loss function to be minimized without decomposition into two sub-problems.\footnote{Note this simplified loss function also drops the term $\left\langle  \lambda, \pmb{\theta} - \textbf{z} \right\rangle$ because $\mathbf{z}$ is updated with $\pmb{\theta}$ at the beginning of each epoch, forcing the Lagrangian multiplier $\lambda$ and its contribution to the loss function to be always 0.} Moreover, SGD with this loss function converges similarly to the SGD with $J(g(\textbf{\em x};\pmb{\theta}, \mathbf{m}), \textbf{\em t})$ and more reliably than ADMM. 
%
%
%
%
Intuitively, the key role of the dynamic regularizer in this simplified loss function 
is to encourage the DNN to not change 
values of the weights that have large magnitude  
unless the corresponding loss is large, similar to what the dynamic regularizer does in ADMM-based pruning. 

\subsection{Proposed Hybrid Loss Function}

For a given input image $\textbf{\em x}$, adversarial training can be viewed as a min-max optimization problem that finds the model parameters $\pmb{\theta}$ that minimize the loss associated with the corresponding adversarial sample $\hat{\textbf{\em x}}$, as shown below:
\begin{equation}\label{eq:min_max}
    \underset{\pmb{\theta}}{\mbox{arg min}}\{{\underset{\mathbf {{\hat {\textbf{\em x}} \in P_{\epsilon}(\textit{\textbf{x}})}}}{\mbox{arg max}} \; {J(g({\hat {\textbf{\em x}}}; \pmb{\theta}),\textbf{\em t})}}\}
\end{equation}
In our framework, we use SGD for loss minimization and PGD to generate adversarial images.
%
More specifically, to boost classification robustness on perturbed data we propose using a hybrid loss function that combines the proposed simplified loss function in Eq. \ref{eq:admm-simplified} with
adversarial image loss, i.e.,
%
\begin{align}\label{eq:tot_loss}
J_{tot} = \beta 
{\mathcal{L}}_{\rho}(\pmb{\theta},\textbf{z}, \mathbf{m})
+ (1-\beta) {{ J(g({\hat {\textbf {\em x}}}; \pmb{\theta},\mathbf{m}),\textbf{\em t})}}
\end{align}
$\beta$ provides a tunable trade-off between the two loss components. 

\textbf{Observation 1} \textit{A DNN only having a fraction of weights active throughout the training can be trained with the proposed hybrid loss to finally converge similar to that of the un-pruned model (mask $\mathbf{m}=1$) to provide a robust yet compressed model}. 

This is exemplified in Fig. \ref{fig:vgg16_c10_loss_sensitivity}(a) which shows similar convergence trends for both pruned and unpruned models, simultaneously achieving both the target compression and robustness while also mitigating the requirement of multiple training iterations.

\subsection{Support for Channel Pruning}

Let the weight tensor of a convolutional layer $l$ be denoted as $\pmb{\theta}_l \in \mathbb{R}^{M \times N \times h \times w}$, where $h$ and $w$ are the height and width of the convolutional kernel, and $M$ and $N$ represent the number of filters and channels per filter, respectively. We convert this tensor to a 2D weight matrix, with $M$ and $N \times h \times w$ being the number of rows and columns, respectively. We  then partition this matrix into $N$ sub-matrices of $M$ rows and $h \times w$ columns, one for each channel. To compute the importance of a channel $c$, we find the Frobenius norm (F-norm) of corresponding sub-matrix, thus effectively compute $O_l^c$ = $\lvert{\pmb{{\theta}}_l^{:,c,:,:}}\rvert^{2}_F$. 
Based on the fraction of non-zero weights that need to be rewired during an epoch $i$, denoted by the pruning rate $p_i$, we compute the number of channels that must be pruned from each layer, ${c}^{p_i}_l$, and prune the ${c}^{p_i}_l$ channels with the lowest F-norms. 
We then compute each layer's importance based on the normalized momentum contributed by its non-zero channels. These importance measures are used to determine the number of zero-F-norm channels $r_l^i \geq 0$ that should be re-grown for each layer $l$. 
More precisely, we re-grow the $r_l^i$ zero-F-norm channels with the highest Frobenius norms of their momentum.
We note that this approach can easily be extended to enable various other forms of structured pruning.
Moreover, despite supporting pruning of both convolution and linear layers, this paper focuses on reducing the computational complexity of a DNN. We thus experiment with pruning only convolutional layers because they dominate the computational complexity  \cite{kundu2020pre}. The detailed pseudo-code of the  proposed training framework is shown in Algorithm \ref{alg:DNR}.
\begin{figure}[!t]
\includegraphics[width=0.48\textwidth]{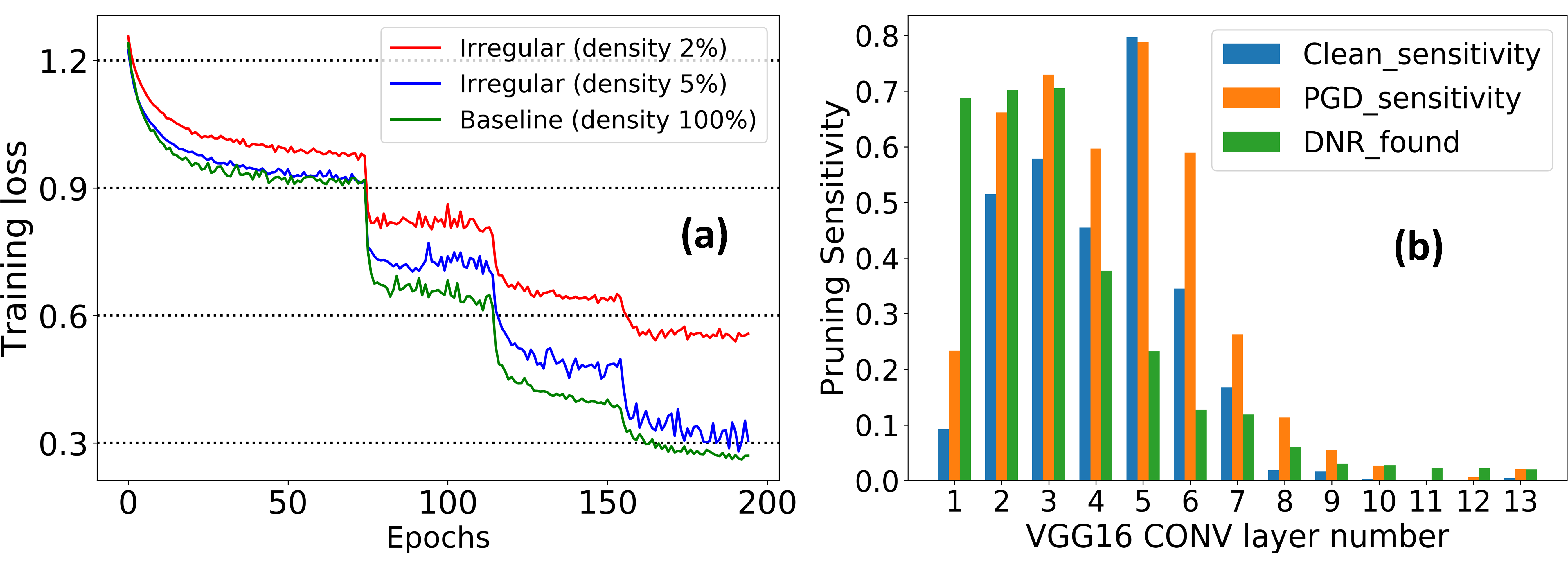}
\centering
\vspace{-0.4cm}
   \caption{(a) Training loss vs. epochs and (b) Pruning sensitivity per layer for VGG16 on CIFAR-10.}
\label{fig:vgg16_c10_loss_sensitivity}
\vspace{-3mm}
\end{figure}
\begin{algorithm}[t]
\small
\SetAlgoLined
\DontPrintSemicolon
\KwData{weight ${\pmb{\theta}}_l$, momentum $\pmb{\mu}_l$, binary mask \textbf{M}$_l, l = 0..k$}
\KwData{density $d$, $ i = 0..\text{numEpochs}$, pruning rate $p = p_{i=0}$ \text{pT: \textbf{irregular} or \textbf{channel}}}
\For{$l \leftarrow 0$ \KwTo $k$}
{
    ${\pmb{\theta}}_l \leftarrow \text{init}({\pmb{\theta}}_l)$\ \&
    $\textbf{M}_l \leftarrow \text{createMaskForWeight}({\pmb{\theta}}_l, d)$\;  
    $\text{applyMaskToWeights}({\pmb{\theta}}_l, \textbf{M}_l)$\;
    ${\bf z}_l \leftarrow {\pmb{\theta}}_l \odot \textbf{M}_l$\;
}

\For{$\text{i} \leftarrow 0$ \KwTo \text{numEpochs}}
{
    \For{$\text{j} \leftarrow 0$ \KwTo \text{numBatches}}
    {
     $\textit{J} = \text{computeCleanLoss(\pmb{\textit x})} + \text{updateDynmicRegularizr(}{\pmb{\theta}, {\bf z})}$\;
     $\textit{J}_{adv} = \text{computePerturbedLoss(}{\hat{\pmb{\textit x}})}$\;
     $\textit{J}_{tot} = \text{updateRobustLoss(}{\textit{J}, \textit{J}_{adv})}$\;
     $\frac{\partial \textit{J}_{tot}}{\partial{\pmb{\theta}}} = \text{computeGradients(}{\pmb{\theta}}, \text{batch}\text{)}$\;
     $\text{updateMomentumAndWeights}(\frac{\partial {\textit{J}}_{tot}}{\partial{\pmb{\theta}}}, \pmb{\mu} )$\;
     \For{$l \leftarrow 0$ \KwTo $k$}
     {
     $\text{applyMaskToWeights}({\pmb{\theta}}_l, \textbf{M}_l)$\;
     }
    }
    $\text{tM} \leftarrow \text{getTotalMomentum}({\pmb{\mu}})$\;
    $\text{pT} \leftarrow \text{getTotalPrunedWeights}({\pmb{\theta}}, p_i)$\;
    $p_i \leftarrow \text{linearDecay}(p_i)$\;
    \For{$l \leftarrow 0$ \KwTo $k$}
    {
        $\pmb{\mu}_l \leftarrow \text{getMomentumContribution}({\pmb{\theta}}_l,
        \textbf{M}_l, \text{tM}, \text{pT})$\;
        $\text{Prune}({\pmb{\theta}}_l, \textbf{M}_l, p_i, \text{pT})$\;
  $\text{Regrow}({\pmb{\theta}}_l, \textbf{M}_l, \pmb{\mu}_l \cdot \text{tP}, \text{pT})$\;
  $\text{applyMaskToWeights}({\pmb{\theta}}_l, \textbf{M}_l)$\;
  ${\bf z}_l \leftarrow {\pmb{\theta}}_l \odot \textbf{M}_l$\;
  }
}
 \caption{DNR Training.}
 \label{alg:DNR}
\end{algorithm}
It is noteworthy that DNR's ability to arrange per-layer pruning ratio for robust compression successfully avoids the tedious task of hand-tuning the pruning-ratio based on \textit{layer sensitivity}. To illustrate this, we follow \cite{ding2019global} to quantify the sensitivity of a layer by measuring the percentage reduction in classification accuracy on both clean and adversarial images caused by pruning that layer by $x\%$ without pruning other layers.

\textbf{Observation 2}\textit{ DNN layers' sensitivity towards clean and perturbed images are not necessarily equal, thus determining layer pruning ratios for robust models is particularly challenging.} 

As exemplified in Fig. \ref{fig:vgg16_c10_loss_sensitivity}(b), for $x$ = $95\%$ there is significant difference in the sensitivity of the layers for clean and perturbed image classification. DNR, on the contrary, automatically finds per-layer pruning ratios (overlaid as pruning sensitivity as in \cite{ding2019global}) that serves well for both types of image classification targeting a global pruning of $95\%$.

\section{Experiments}
\label{sec:res}
Here, we first describe the experimental setup we used to evaluate the effectiveness of the
proposed robust training scheme.
We then compare our
method against other state-of-the-art robust pruning techniques based on ADMM \cite{ye2019adversarial} and $L_1$ lasso \cite{rakin2019robust}. We also evaluate the merit of DNR as a clean-image pruning scheme and show that it consistently outperforms contemporary non-iterative model pruning techniques  \cite{lee2018snip, dettmers2019sparse, ding2019global, He_2018_ECCV}. We finally present an ablation study to empirically evaluate the importance of the dynamic regularizer in the DNR's loss function. 
We used Pytorch \cite{paszke2017automatic} to write the models and trained/tested on AWS P3.2x  large instances with an NVIDIA Tesla V100 GPU.  
\subsection{Experimental Setup}\label{subsec:setup}
\subsubsection{Models and Datasets}
We selected three widely used datasets, CIFAR-10 \cite{krizhevsky2009learning} CIFAR-100 \cite{krizhevsky2009learning} and Tiny-ImageNet \cite{hansen2015tiny} and picked two well known CNN models, VGG16 \cite{simonyan2014very} and ResNet18 \cite{he2016deep}. Both CIFAR-10 and CIFAR-100 datasets have 50K training samples and 10K test samples with an input image size of $32 \times 32 \times 3$. Training and test data size for Tiny-ImageNet are 100k and 10k, respectively where each image size is of $64 \times 64 \times 3$. For all the datasets we used standard data augmentations (horizontal flip and random crop with reflective padding) to train the models with  a batch size of 128. 
\subsubsection{Adversarial Attack and DNR Training Settings}
For PGD, we set $\epsilon$ to $8/255$, the attack step size  $\alpha = 0.01$, and the number of attack iterations to $7$, the same values as in \cite{he2019parametric}. 
For FGSM, we choose the same $\epsilon$ value as above. 
 
 We performed DNR based training for 200/170/60 epochs for CIFAR-10/CIFAR-100/Tiny-ImageNet, with a starting learning rate of $0.1$, momentum value of $0.9$, and weight decay value of $5e^{-4}$. For CIFAR-10 and CIFAR-100 the learning rate (LR) was reduced by a factor of $0.2$ after $80$, $120$, and $160$ epochs. For Tiny-ImageNet we reduced the LR value after $30$ and $50$ epochs. In addition, we hand-tuned $\rho$ to $10^{-4}$ and set the pruning rate $p=0.5$. We linearly decreased the pruning rate every epoch by $\frac{p}{total\; epochs}$. Finally, to balance between the clean and adversarial loss, we set $\beta$ to $0.5$. Lastly, note that we performed warm-up sparse learning \cite{dettmers2019sparse} for the first 5 epochs with only the clean image loss function before using the hybrid loss function with dynamic regularization (see Eq. \ref{eq:tot_loss}) for robust compression for the remaining epochs.  
\subsection{Results}\label{subsec:res}
\begin{figure*}[t]
\includegraphics[width=0.95\linewidth]{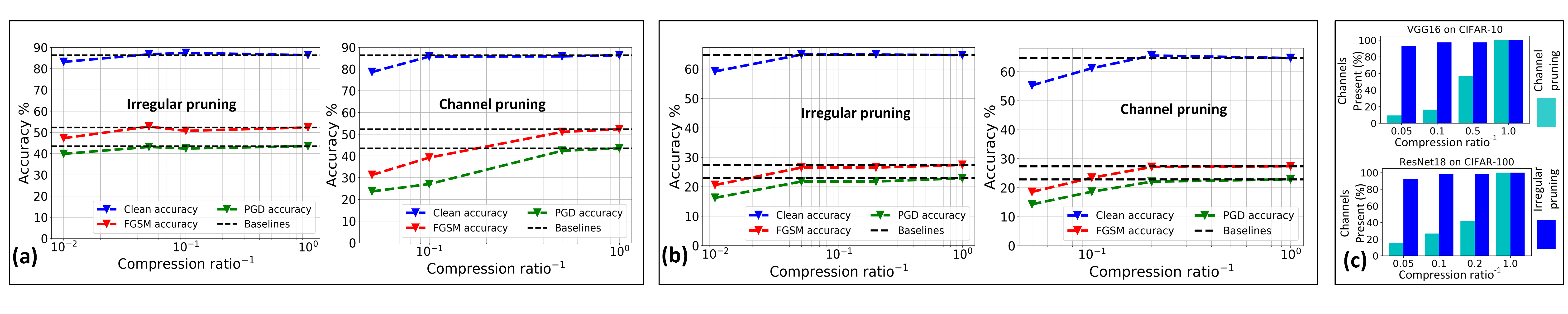}
\centering
\vspace{-2mm}
   \caption{ Model compression vs. accuracy (on both clean and adversarially generated images) for irregular and channel pruning evaluated with (a) VGG16 on CIFAR-10 and (b) ResNet18 on CIFAR-100. (c) Comparison of channel pruning with irregular pruning in terms of \% of channels present. Note that the \% of channels present correlates with inference time \cite{liu2018rethinking, dettmers2019sparse}.}
\label{fig:vgg_c10_res_c100}
\end{figure*}
\textbf{Results on CIFAR datasets:} We analyzed the impact of our robust training framework on both clean and adversarially generated images with various target compression ratios in the range $[0.01, 1.0]$, where
model compression is computed as the ratio of total weights of the model to the non-zero weights in the pruned model.
As shown in Figs.~\ref{fig:vgg_c10_res_c100}(a-b) DNR can effectively find a robust model with high compression and negligible compromise in accuracy. In particular, for irregular pruning our method can compress up to $\mathord{\sim}20\times$ with negligible drop in accuracy on clean as well as PGD and FGSM based perturbed images, compared to the baseline non-pruned models, tested with VGG16 on CIFAR-10 and  ResNet18 on CIFAR-100.\footnote{A similar trend is observed for VGG16 on CIFAR-100 and ResNet18 on CIFAR-10. These are not included in the paper due to space limitations.}

\textbf{Observation 3}\textit{ As the target compression ratio increases, channel pruning degrades adversarial robustness more significantly than irregular pruning}. 

As we can see in Fig. \ref{fig:vgg_c10_res_c100}(a-b), the achievable model compression with negligible accuracy loss for structured (channel) pruned models is $\mathord{\sim}10\times$ lower than that achievable through irregular pruning. This trend matches with that of the model's performance on clean image. However, as we can see in Fig. \ref{fig:vgg_c10_res_c100}(c), the percentage of channels present in our channel-pruned models can be up to $\mathord{\sim}10\times$ lower than its irregular counterparts, implying a similarly large speedup in inference time on a large range of compute platforms \cite{dettmers2019sparse}.
\newline
\textbf{Results on Tiny-ImageNet}: As shown in Table \ref{tab:tiny_robust}, DNR can compress the model up to $20.63\times$ without any compromise in performance for both clean and perturbed image classification. 

It is also noteworthy that all our accuracy results for both clean and adversarial images correspond to models that provide the best test accuracy on clean images. This is because robustness gains are typically more relevant on models in which the performance on clean images is least affected. 
\begin{table}
\begin{center}
\scriptsize\addtolength{\tabcolsep}{-2.5pt}
\begin{tabular}{|c|c|c|c|c|c|}
\hline
Pruning & Compression & \% Channel & \multicolumn{3}{|c|}{Accuracy (\%)}\\
\cline{4-6}
type  &  -ratio & present & Clean   & FGSM & PGD  \\
\hline
Unpruned-baseline & $1\times$ & 100 & 50.91 & 18.19 & 13.87 \\
Irregular & $\textbf{20.63}\times$ & 98.52 & \textbf{51.71} & \textbf{18.21} & \textbf{14.46} \\
Channel & $1.45\times$ & \textbf{74} & 51.09 & 17.92 & 13.54 \\
\hline
\end{tabular}
\end{center}
\caption{Results on VGG16 to classify Tiny-ImageNet.}
\label{tab:tiny_robust}
\vspace{-2mm}
\end{table}
\subsection{Comparison with State-of-the-art}
Here, were compare the performance of DNR with ADMM \cite{ye2019adversarial} and $L_1$ lasso based \cite{rakin2019robust} robust pruning. For ADMM based robust pruning we followed a three stage compression technique namely {\em pre-training}, {\em ADMM based pruning}, and {\em masked retraining}, performing pruning for 30 epochs with $\rho_{admm} = 10^{-3}$ as described in \cite{ye2019adversarial}. $L_1$ lasso based pruning  adds a $L_1$ regularizer to its loss function to penalize the weight magnitudes, where the regularizer coefficient determines the penalty factor. Table \ref{tab:dnr_robust_comp} shows that our proposed method outperforms both ADMM and $L_1$ Lasso based approaches by a considerable margin, retaining advantages of both worlds \footnotemark. In particular, compared to ADMM, with VGG16 (ResNet18)  model on CIFAR-10, DNR provides up to $3.4\%$ ($0.78\%$) increased classification accuracy on perturbed images with $1.24\times$ ($1.48\times$) higher compression. Compared to $L_1$ Lasso, we achieve $10.38\times$ ($3.15\times$) higher compression and up to $2.6\%$ ($0.55\%$), and $3.5\%$ ($1.4\%$) increased accuracy on perturbed and clean images, respectively, for VGG16  (ResNet18) on CIFAR-10 classification.
\footnotetext{Romanized numbers in the table are results of our experiments, and italicized values are directly taken from the respective original papers.}

\textbf{Observation 4}\textit{ Naively tuned per-layer pruning ratio degrades both robustness and clean-image classification performance of a model.} 

For this, we evaluated robust compression using naive ADMM, i.e. using naively tuned per-layer pruning-ratio (all but the 1st layer $\mathord{\sim}x\%$ for a $x\%$ total sparsity). As shown in Table \ref{tab:dnr_robust_comp}, this clearly degrades the performance, implying layer-sparsity tuning is necessary for ADMM to perform well.

\begin{table}
\begin{center}
\scriptsize\addtolength{\tabcolsep}{-6.5pt}
\begin{tabular}{|c|c|c|c|c|c|c|c|c|c|}
\hline
{} & {} & No pre- & Per-layer & Target  & Pruning & Compre-  & \multicolumn{3}{|c|}{Accuracy (\%)}\\
\cline{8-10}
Model &  Method & trained   & sparsity & pruning & type  & ssion & {} &  {} & {} \\
{}  &  {}   &  model & knowledge & met & {}  & ratio & Clean &  FGSM & PGD \\
{}  &  {}   &  {} & not-needed & {} & {}  & {} & {} &  {} & {} \\
\hline\hline
   & ADMM \cite{ye2019adversarial} & $\times$ & $\times$ & \checkmark    & Irregular & $16.78\times$ & $86.34$ & $49.52$ & $40.62$\\
VGG16      & ADMM naive & $\times$ & \checkmark & \checkmark &
      & $19.74\times$ & 83.87 & 42.46 & 32.87 \\
       & $L_1$ Lasso \cite{rakin2019robust} &  \checkmark &  \checkmark  & $\times$ &  & $\textit{2.01}\times$ & $\textit{83.24}$ & $\textit{50.32}$ & $\textit{42.01}$\\
      & DNR   & \checkmark & \checkmark  &  \checkmark    &           & $\textbf{20.85}\times$ & $\textbf{86.74}$ & $\textbf{52.92}$ & $\textbf{43.21}$\\
\hline
 & ADMM \cite{ye2019adversarial} & $\times$ & $\times$ & \checkmark      & Irregular & $14.6\times$ & $87.15$ & $54.65$ & $46.57$ \\
ResNet18 & ADMM naive & $\times$ & \checkmark & \checkmark &
      & $19.74\times$ & 86.10 & 50.49 & 42.24 \\
      & $L_1$ Lasso \cite{rakin2019robust} & \checkmark &  \checkmark & $\times$   &           & $\textit{6.84}\times$ & $\textit{85.92}$ & $\textbf{\em 55.20}$ & $\textit{46.80}$\\
     & DNR  & \checkmark & \checkmark &  \checkmark      &           & $\textbf{21.57}\times$ & $\textbf{87.32}$ & $55.13$ & $\textbf{47.35}$\\
\hline
\end{tabular}
\end{center}
\caption{Comparison of DNR, ADMM based, and $L_1$ lasso based robust pruning schemes on CIFAR-10.}
\label{tab:dnr_robust_comp}
\vspace{-1mm}
\end{table}
\subsection{Pruning to Classify Clean-only Images}
\label{subsec:dnr_clean}
To evaluate the merit of DNR as a clean-image only pruning scheme (DNR-C), we trained using DNR with the same loss function minus the adversarial loss term (by setting $\beta = 1.0$ in Eq. \ref{eq:tot_loss}) to reach a target pruning ratio. Table \ref{tab:dnr_pruning_only} shows that our approach consistently outperforms other state-of-the-art non-iterative pruning approaches based on momentum information \cite{ding2019global, dettmers2019sparse}, reinforcement-learning driven auto-compression (AMC) \cite{He_2018_ECCV}, and connection-sensitivity \cite{lee2018snip}\footnotemark[\value{footnote}]. 
The $\delta$ value in the seventh column represents the error difference from corresponding non-pruned baseline models. We also present performance on CIFAR-100 for VGG16 and ResNet18 and Tiny-ImageNet for VGG16.\footnote[6]{To have an "apple to apple" comparison we provide results on ResNet50 model for classification on CIFAR-10. All other simulations are done on only the ResNet18 variant of ResNet.} In particular, we can achieve up to $34.57\times$ ($12.61\times$) compression on CIFAR-10 dataset with irregular (channel) pruning maintaining accuracy similar to the baseline. On CIFAR-100 compression of up to $22.45\times$ ($5.57\times$) yields no significant accuracy drop (less than $2.7\%$ in top-1 accuracy) with irregular (channel) pruning. Moreover, our evaluation shows a possible practical speed up of up to $6.06\times$ for CIFAR-10 and $2.41\times$ for CIFAR-100 can be achieved through channel pruning using DNR-C. For Tiny-ImageNet, DNR-C can provide compression and speed-up of up  to $11.55\times$ and $1.53\times$, respectively with negligible accuracy drop.

\begin{table}
\scriptsize\addtolength{\tabcolsep}{-6.2pt}
\begin{center}
\begin{tabular}{|c|c|c|c|c|c|c|c|}
\hline
Dataset & Model & Method & Pruning & Compress-  & {Error (\%)} & $\delta$ from & Speedup\\

{} & {}  &  {}   & type  & ion ratio & top-1  &  baseline & \\
\hline\hline
{} & VGG16   & SNIP \cite{lee2018snip}     & Irregular & $\textit{32.33}\times$ & $\textit{8.00}$ & -0.26 & --\\
 &       & Sparse-learning \cite{dettmers2019sparse}  &           & $\textit{32.33}\times$ & $\textit{7.00}$ & -0.5 & --\\
  &    & DNR-C       &           & $\textbf{34.57}\times$ & $\textbf{6.50}$ & -0.09 & $1.29\times$\\
\cline{4-8}
   &    & DNR-C       &    Channel       & $12.61\times$ & $8.00$ & -1.5 & $6.06\times$\\
\cline{2-8}
CIFAR  & ResNet50 
  & GSM  \cite{ding2019global} & Irregular & $\textit{10}\times$ & $\textit{6.20}$ & $\textit{-0.25}$ & --\\
-10 &      & AMC \cite{He_2018_ECCV}      &           & $\textit{2.5}\times$ & $\textit{6.45}$ & $\textit{+0.02}$ & --\\
 &      & DNR-C                       &           & $\textbf{20}\times$ & $\textbf{4.8}$ & -0.07 & $1.75\times$ \\
\cline{2-8}
 & ResNet18   & DNR-C & Irregular & $20.32\times$ & $5.19$ & -0.10 & $1.31\times$\\
\cline{4-8}
 & {}   &      & Channel & $5.67\times$ & $5.36$ & -0.27 & $2.43\times$\\
\hline
\hline
 & VGG16 & DNR-C & Irregular & $20\times$ & 27.14 & -1.04 &  $1.07\times$ \\
\cline{4-8}
CIFAR &  &  & Channel & $2.76 \times$ & 28.78 & -2.68  & $2.06\times$\\
\cline{2-8}
-100 & ResNet18 & DNR-C & Irregular & $22.45\times$ & 24.9 & -1.17 & $1.13\times$ \\
\cline{4-8}
     &  &  & Channel & $5.57\times$ & $25.28$ & -1.55 & $2.41\times$ \\
\hline
\hline
 Tiny & VGG16 & DNR-C & Irregular & $11.55\times$ & 40.96 & +0.36 &  $1.01\times$ \\
\cline{4-8}
ImageNet &  &  & Channel & $1.74 \times$ & 42.61 & -1.28  & $1.53\times$\\
\hline
\end{tabular}
\end{center}
\caption{Comparison with state-of-the-art non-iterative pruning schemes on CIFAR-10 and comparison of deviation from baseline on CIFAR-100 and Tiny-ImageNet.}
\label{tab:dnr_pruning_only}
\vspace{-2mm}
\end{table}

\begin{table}[]
\scriptsize\addtolength{\tabcolsep}{-3.5pt}
\begin{center}
\begin{tabular}{|c|c|c|c|c||c|c|c|}
\hline
{} & {} & \multicolumn{3}{|c||}{Accuracy (\%) with} & \multicolumn{3}{|c|}{Accuracy (\%) with} \\
Model & Method: DNR & \multicolumn{3}{|c||}{irregular pruning} & \multicolumn{3}{|c|}{channel pruning} \\
\cline{3-8}
{} & {} & Clean & FGSM & PGD & Clean  & FGSM & PGD \\
\hline
\hline
VGG16 & Without dynamic $L_2$ & $\textbf{87.01}$ & 50.09 & 40.62 & $\textbf{86.28}$ & 49.49 & 41.25 \\
{} & With dynamic $L_2$ & 86.74 & $\textbf{52.92}$ & $\textbf{43.21}$ & 85.83 & $\textbf{51.03}$ & $\textbf{42.36}$ \\
\hline
\hline
ResNet18 & Without dynamic $L_2$ & $\textbf{87.45}$ & 53.52 & 45.33 & $\textbf{87.97}$ & 53.10 & 45.91 \\
{} & With dynamic $L_2$ & 87.32 & $\textbf{55.13}$ & $\textbf{47.35}$ & 87.49 & $\textbf{56.09}$ & $\textbf{48.33}$ \\
\hline
\end{tabular}
\end{center}
\caption{Comparison of DNR with and without the dynamic regularizer for CIFAR-10 classification.}
\label{tab:dnr_ablation}
\vspace{-4mm}
\end{table}
\subsection{Ablation Study}
\label{subsec:abl}
To understand the performance of the proposed hybrid loss function with a dynamic $L_2$ regularizer, we performed ablation with both VGG16 and ResNet18 on CIFAR-10 for a target parameter density of $5\%$ and $50\%$ using irregular and channel pruning, respectively. As shown in Table \ref{tab:dnr_ablation}, using the dynamic regularizer improves the adversarial classification accuracy by up to $2.83\%$ for VGG16 and $\mathord{\sim}3\%$ for ResNet18 with similar clean-image classification performance.

\subsection{Generalized Robustness Against PGD Attack of Different Strengths}
\begin{figure}
\includegraphics[width=0.80\linewidth]{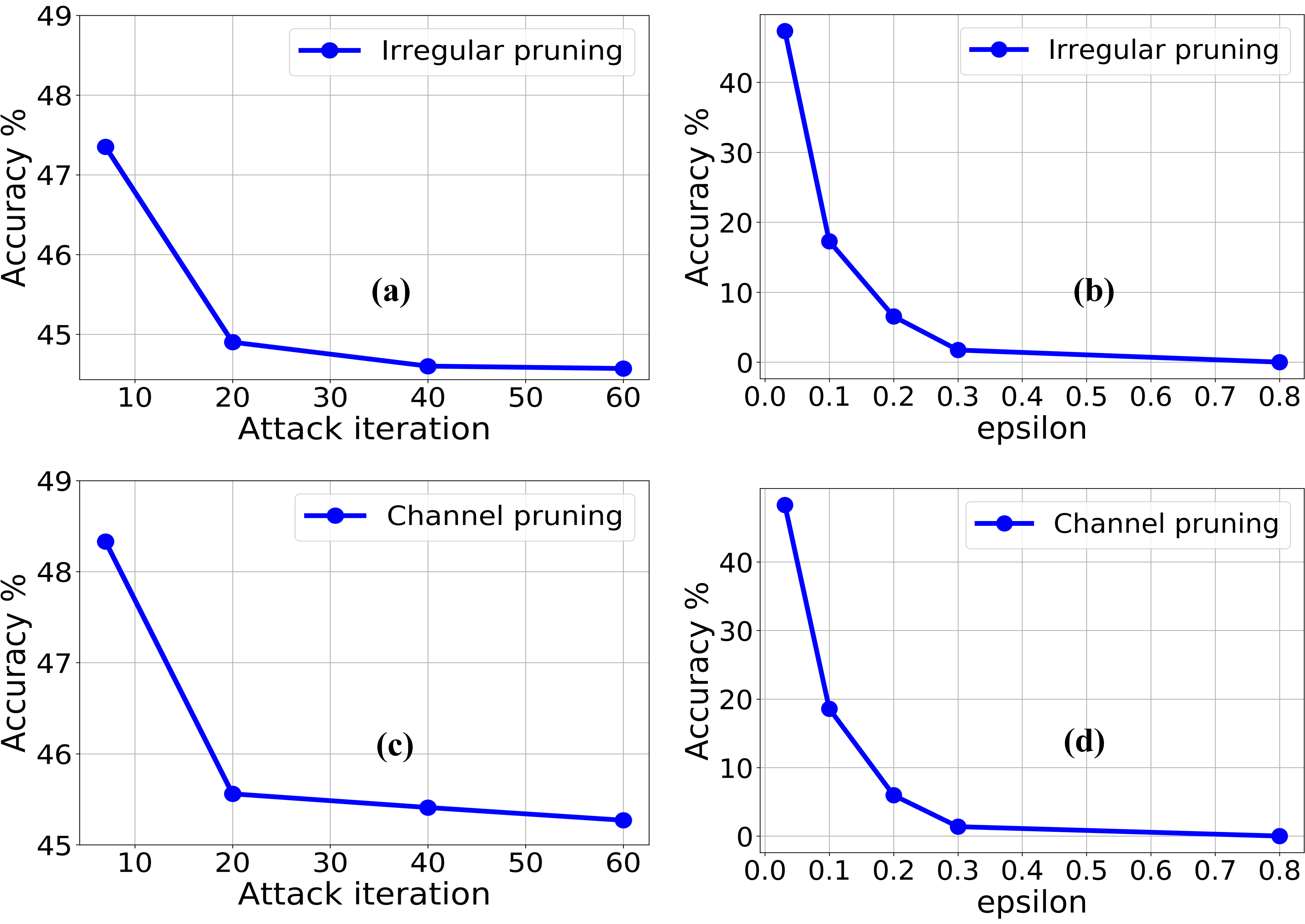}
\centering
   \caption{On CIFAR-10, the perturbed data accuracy of ResNet18 under PGD attack versus increasing  (a), (c) attack iteration and (b), (d) attack bound $\epsilon$ for irregular ($5\%$ density), and channel pruned ($50\%$ density) models, respectively.}
\label{fig:res18_c10_iter_ep_attack_strength}
\end{figure}

Fig. \ref{fig:res18_c10_iter_ep_attack_strength} presents the performance of the pruned models as a function of the PGD attack iteration and the attack bound $\epsilon$. In particular, we can see that, for both irregular and channel pruned models, the accuracy degrades with higher number of attack iterations. When $\epsilon$ increases, the accuracy drop is similar in both the pruning schemes. These trends suggest that our robustness is not achieved via gradient obfuscation \cite{rakin2019robust}.

\section{Conclusions}
\label{sec:con}
This paper addresses the open problem of achieving ultra-high compression of DNN models while maintaining their robustness through a  non-iterative training approach. In particular, the proposed DNR method leverages a novel sparse-learning strategy with a hybrid loss function that has a dynamic regularizer to achieve better trade-offs between accuracy, model size, and robustness. Furthermore, our extension to support channel pruning shows that compressed models produced by DNR can have a practical inference speed-up of up to $\mathord{\sim}10\times$.
\bibliographystyle{IEEEbib}
\bibliography{refs}

\end{document}